# An ASR-based tutor for learning to read: how to optimize feedback to first graders

Yu Bai[1], Cristian Tejedor-García[1,2], Ferdy Hubers[1,2], Catia Cucchiarini[1], and Helmer Strik[1,2,3,4]

[1] Centre for Language and Speech Technology (CLST), Radboud University Nijmegen
[2] Centre for Language Studies (CLS), Radboud University Nijmegen
[3] Donders Institute for Brain, Cognition and Behaviour, Radboud University Nijmegen
[4] NovoLearning B.V., Nijmegen
y.bai@let.ru.nl

**Abstract.** The interest in employing automatic speech recognition (ASR) in applications for reading practice has been growing in recent years. In a previous study, we presented an ASR-based Dutch reading tutor application that was developed to provide instantaneous feedback to first-graders learning to read. We saw that ASR has potential at this stage of the reading process, as the results suggested that pupils made progress in reading accuracy and fluency by using the software. In the current study, we used children's speech from an existing corpus (JASMIN) to develop two new ASR systems, and compared the results to those of the previous study. We analyze correct/incorrect classification of the ASR systems using human transcripts at word level, by means of evaluation measures such as Cohen's Kappa, Matthews Correlation Coefficient (MCC), precision, recall and F-measures. We observe improvements for the newly developed ASR systems regarding the agreement with human-based judgment and correct rejection (CR). The accuracy of the ASR systems varies for different reading tasks and word types. Our results suggest that, in the current configuration, it is difficult to classify isolated words. We discuss these results, possible ways to improve our systems and avenues for future research.

**Keywords:** Automatic speech recognition · ASR · reading tutor · child speech

## 1 Introduction

The idea that automatic speech recognition (ASR) technology can be employed to support children learning to read has been around for quite a while [26, 36, 32]. Various research systems and commercial products have been proposed [20, 27, 11] that can monitor children while reading and indicate when they encounter difficulties. Fewer ASR-based systems are really capable of providing a kind of guidance and feedback that is comparable to what teachers usually offer in their classrooms. Besides, there is still scarce empirical evidence about the effectiveness of ASR-based systems for Dutch child speech. In the project Dutch Automatic Reading Tutor (DART[5]) we investigate whether such a system can be developed for the initial stages of learning to read (grade 1) when children need to develop accurate and rapid word decoding skills [17, 28]. In this paper we report on experiments that were conducted to investigate the performance of different ASR systems in identifying reading errors and to provide feedback on those errors.

To improve the ASR system in reading tutors, evaluating ASR miscue detection is important. A good reading tutor for children should avoid too many false rejections (FRs), which will lower the credibility of the reading tutor and cause frustration, and detect miscues that pupils can correct [25]. In the previous study, we evaluated the agreement between DART reading tutor with Novo-ASR backend and human transcripts [3]. The highest agreement obtained measured by Cohen's Kappa is .41, which is classified as moderate agreement [18]. In this study, we first developed two ASR systems based on Dutch child speech and investigated the performance of three ASR systems in respect of how accurate the ASR judgments are compared to human judgments. For this purpose we employed various metrics, including Cohen's Kappa, Matthews Correlation Coefficient (MCC), precision, recall and F-measures. The material used in the reading tutor contains four reading tasks designed with different goals and difficulty levels for first graders [2]. We also investigated the ASR accuracy in each reading task and different word types from two reading tasks which contain both words that are familiar to the first graders and new words.

The rest of the paper is organized as follows. Section 2 describes the research background of educational applications aiming at helping children improve reading skills and Kaldi as a framework for ASR technology. Section 3 formulates the aim of the study and describes the reading tutor. Section 4 explains the data used in the experiments, how the new ASR systems were built with kaldi and the procedure of how the three ASR systems were evaluated. The results of performance evaluation of the ASR systems are reported in section 5. Finally, in section 6 we discuss the results, possible ways of improving our systems and new lines of future work.

## 2   Research Background

Most of the research that has been conducted so far on applying ASR in the context of reading skills acquisition [12, 31, 36] addressed reading skills in English [25, 26, 32]. Many of the studies addressed reading assessment at a global level [35, 5] or were aimed at monitoring children while they read aloud whole passages and at providing some form of support when they encounter difficulties [9]. For English even several commercial products have been made available for this latter purpose, such as the Reading Assistant[6], IBM Reading Companion[7], and the ReadingBuddy[8]. A possible application of ASR could be at earlier stages of the process of learning to read, when children need to acquire grapheme-phoneme connections in order to develop word decoding skills [7]. This process requires intensive practice in reading aloud that cannot always be provided in teacher-fronted education. So far, relatively few studies have investigated to what extent ASR can be employed in this type of application. This is not surprising, as such applications are extremely challenging in terms of ASR performance and accuracy.

---

[5] http://hstrik.ruhosting.nl/DART/

Feedback provided by such reading tutor applications is not always correct. Various methods of evaluating ASR-based reading tutoring systems have been discussed. Taking human classification as the criterion, agreement (Cohen's Kappa value) of automatic classifiers turns out to vary for reading tasks with different numbers of syllables [9]. The agreement rate of feedback messages between system judgment and human judgment was used as a measure [1]. Miscue detection rate and false alarm rate can also be used as evaluation metrics [4]. Since most reading tutor applications are task-oriented, their performance can differ from one type of task to another.

Regarding existing ASR technology, Kaldi is one of the most current and state-of-the-art open-source toolkits [29], becoming a highly matured development tool for almost any language in the speech recognition area thanks to its recipe scripts. Kaldi uses a finite-state transducer (FST)-based framework; therefore, any language model can be applied which supports FST. One can easily implement N-gram language model using the SRILM toolkit [19]. Recent research studies integrated Kaldi for assessing the pronunciation of non-native speakers producing non-spontaneous speech [33, 34] and spontaneous speech [14, 30]. In the case of reading assessment one of the main problems is dealing with child speech [9, 5]. In the specific case of first graders this is even more problematic because these children are in the early stages of the reading process and therefore produce many hesitations, disfluencies and broken words. Addressing these difficulties and providing the appropriate feedback online, makes this task extremely challenging

## 3 Present Study

### 3.1 Aim of the Study

Developing reading skills is crucial for first-grade pupils. Reading software systems that incorporate ASR are able to provide more opportunities for pupils to practice reading aloud and receiving immediate feedback. In the DART project, we aim to investigate how such reading tutor software with ASR can be best developed and implemented. We especially focus on feasible and suitable forms of feedback on reading accuracy and fluency.

---

[6] http://www.readingassistant.com/
[7] https://www.ibm.com/ibm/responsibility/downloads/initiatives/ReadingCompanion.pdf
[8] http://readingbuddysoftware.com/

Reading accuracy is an important reading skill and providing appropriate feedback on reading accuracy is crucial in a reading tutor. The forms of feedback depend on the correct/incorrect judgment by the system based on the output of the ASR. The aim of the present study is to gain insight into the accuracy of binary classification based on the outputs of different ASR systems and how to improve them. We investigate to what extent a) using ASR trained on child speech helps improve the agreement with human judgment, b) to what extent the improvement differs in different reading tasks and word types and c) to what extent ASR performance varies in terms of miscue detection and false rejection.

### 3.2  The Reading Tutor

The reading tutor is based on a reading method for first graders developed by Zwijsen Publishers, 'Veilig Leren Lezen'. It distinguishes two types of exercises: accuracy exercises and fluency exercises and four reading tasks: reading a) isolated words b) sentences c) word lists and d) stories. The accuracy exercises are meant to enhance pupils' reading accuracy of individual words and sentences. Sentences and isolated words are read one by one and each word is analysed by an ASR backend. Based on this analysis, pupils receive feedback on whether the target word or sentence is read correctly. If incorrect, the pupil has to try again (up to three attempts). The fluency exercises aim to improve the pupils' reading fluency and accuracy. In these exercises, pupils read word lists or stories in one go. Subsequently, they receive feedback and have to try the incorrect words or sentences again. Next, pupils have to read the same word list or story again. For more information about the system and the reading materials see [2].

## 4  Method

### 4.1  Transcribed Speech Data

In 2020, the first experiment of the DART project was conducted with school children practicing from home because of the pandemic. In spite of the complexities in organising this experiment, we managed to collect 1547 audio recordings which were subsequently transcribed by an annotator using Praat [3]. After filtering recordings that, according to the annotator, were either empty or damaged, or contained excessive background noise, 1132 audio files with 8263 words were left. These data were used to test the accuracy of three ASR systems we compared.

### 4.2  Kaldi ASR Systems Pipeline

The machine configuration on which the experiment was conducted has Ubuntu 18.04.1 LTS (64-bit operating system), AMD EPYC 7502P 32-Core (64 threads) processor with 2.5-3.35 GHz. We installed Kaldi on the machine. We adopted a classic hybrid training and decoding framework using a simple deep neural network (DNN) with hyperbolic tangent (tanh) nonlinearities [14] after training a context-dependent triphone system with a simple GMM–HMM model [23]. The features are Mel frequency cepstral coefficients (MFCCs) with per-speaker cepstral mean and variance statistics. Since Kaldi underlies on an FST-based framework to build language models from the raw text, we used the SRILM toolkit for building a 4-gram language model [24].

First, we trained monophone models. The first triphone pass used MFCCs, delta and delta–delta features. The second triphone pass included linear discriminant analysis (LDA) and Maximum Likelihood Linear Transform (MLLT). The third triphone pass combined LDA, MLLT and speaker adaptive training (SAT). The next steps included standard Gaussian mixture models (SGMM2), maximum mutual information (MMI) and boosted MMI on top of LDA and MLLT. Finally, we trained the hybrid DNN. For every step mentioned above we applied the appropriate transforms, alignments and decoding techniques for Kaldi[9].

### 4.3   ASR Systems

The reading tutor used by pupils during the experiments made use of the NovoLearning ASR engine, in which the ASR model (ASR.1) was trained with Dutch adult speech. It calculates probability scores and gives right/wrong feedback according to the adjustable threshold [3]. Two other systems (system ASR.2 and system ASR.3) were built with Kaldi after the experimental tests with children. Both systems employed the same acoustic models trained on 7.5 effective hours of child speech material from the JASMIN corpus [8]. The speakers are from JASMIN's group 1: native children between 7 and 11 years old. Different language models were used in these two models. A 4-gram language model generated from a lexicon of prompts was used in model ASR.2, obtaining a word error rate (WER) value of 38.78% (95% confidence interval [35.18, 42.37]); while a 4-gram language model created from a lexicon of orthographic transcriptions was used in model ASR.3, obtaining a WER of 35.71% (95% confidence interval [32.84, 38.58]).

### 4.4   Procedure for Performance Evaluation of ASR Systems

The annotator's transcription was compared to the prompt using the dynamic programming (alignment) algorithm ADAPT [15]. This yielded the reference binary classification word scores (0 for incorrect, 1 for correct). See [3] for more information on this procedure. Using the same procedure, the outputs for the ASR systems 2 and 3 (i.e., the recognized words) are aligned with the prompt, yielding the binary classification scores for these ASR systems. Finally, the later

---

[9] http://kaldi-asr.org/doc/

ASR binary scores are compared to the reference binary scores (which are based on the annotator's transcription).

For ASR.1 the evaluation procedure is different. The output of ASR.1 contains the words in the prompt with probability measures at word level. These probability measures are compared to a threshold to obtain classification scores. In a previous paper [3], we focused on optimizing this threshold.

Two agreement measures were used to evaluate the classification performance. Unbalanced distributions of binary scores were observed in the human annotation, outputs of ASR.1, ASR.2 and ASR.3, which is understandable as pupils generally read more correct words than incorrect words. This is especially the case in Dutch education, in which reading methods are based on incrementally difficult reading levels and children are encouraged to practice words that they are supposed to be capable of reading [21]. A more suitable measure for binary classification in the case of unbalanced data is the MCC [6]. Therefore, we calculated both Cohen's Kappa and MCC as agreement measures for our data. To capture an essence of ASR performance where CA, CR, FA, FR are important, precision, recall and F-measures were also calculated.

Function words, in general have little lexical meaning and mainly express grammatical relation. They are, on average, shorter, are articulated less clearly [10], occur more frequently, and are more familiar to first-grade pupils. Content words possess more semantic content, are less frequent, on average, and often require more reading practice. It is thus interesting to compare the performance of the ASR systems on function vs. content words. To gain insights, agreement measures (Cohen's Kappa and MCC) were calculated for the function words and content words respectively in two reading tasks containing both function words and content words: reading sentences in accuracy exercises and reading stories in fluency exercises.

## 5 Results

In this section we present an evaluation of three different ASR systems using different performance measures, including Cohen's Kappa, MCC, precision, recall and F-measure. Performance on different types of exercises and comparison between agreement on function words and content words are also presented.

### 5.1 Performance Evaluation of Different Systems

In the previous study [3] we found that 48 was the threshold that produced the highest agreement between the ASR scores for correct/incorrect and the human judgment [3]. After removing invalid audio recordings, the best threshold shifted to 46. Table 1 illustrates that the highest Cohen's kappa value of ASR.1 is .42 at a threshold of 46. The models trained on JASMIN native children's speech show better agreement with the human-based judgments with Cohen's kappa of .44 for ASR.2 and .53 for ASR.3.

Table 1 also indicates that model ASR.1 has the lowest MCC value of .42. ASR.2 shows slightly better agreement with human-based judgment than ASR.1 with an MCC of .46. Among three models, ASR.3 shows the highest agreement. This is consistent with Cohen's Kappa values.

**Table 1.** Cohen's Kappa and MCC values of three ASR systems.

| Systems | Cohen's Kappa | MCC |
|---|---|---|
| ASR.1 | .422 | .423 |
| ASR.2 | .442 | .455 |
| ASR.3 | .533 | .568 |

**Table 2.** Correct acceptance rate (CAR), correct rejection rate (CRR), false acceptance rate (FAR), false rejection rate (FRR), precision, recall and F-measure of correct acceptance (CA) and correct rejection (CR).

| | Percentage (%) | | | | Precision (%) | | Recall (%) | | F-measure (%) | |
|---|---|---|---|---|---|---|---|---|---|---|
| | CA | CR | FA | FR | CA | CR | CA | CR | CA | CR |
| ASR.1 | 72.4 | 9.9 | 7.6 | 10.0 | 90.5 | 49.8 | 87.8 | 56.6 | 89.1 | 53.0 |
| ASR.2 | 68.9 | 12.1 | 5.5 | 13.6 | 92.6 | 47.0 | 83.5 | 68.7 | 87.8 | 55.8 |
| ASR.3 | 67.3 | 15.3 | 2.3 | 15.1 | 96.7 | 50.2 | 81.6 | 87.1 | 88.6 | 63.7 |

Table 2 demonstrates correct acceptance (CA), correct rejection (CR), false acceptance (FA), false rejection (FR) rates, precision, recall and F-measure of CA and CR. It shows that there are more CRs and FRs given by the judgment of ASR.2 and ASR.3 while fewer CAs and false acceptances FAs, compared to ASR.1. We see relatively balanced FA and FR in the judgement of ASR.1. Model ASR.3 shows a significant improvement in recall of CR, from 56.6% to 87.10%, but a decrease in recall of CA from 87.85% to 81.64%. In general, all three models show better results for CA, while the precision of CR and F-measure of CR are relatively low. A significant difference in Recall of CR is observed between the three systems.

### 5.2   Performance on Different Reading Tasks

Cohen's Kappa and MCC are calculated for four different reading tasks. Table 3 suggests that all three models perform worse for isolated words than for sentences, word lists and stories. Cohen's Kappa and MCC values of sentences, word lists and stories are similar. However, the values for isolated words are between .20 and .41. The primary cause of the low agreement for isolated words is due to a consequence of less effective language models for isolated words. After removing isolated words from the data, both Cohen's Kappa and MCC are improved (see Table 4).

**Table 3.** Cohen's Kappa and MCC values of three ASR systems for different reading tasks.

|       | Cohen's Kappa | | | | MCC | | | |
|---|---|---|---|---|---|---|---|---|
|       | isolated word | sentence | word list | story | isolated word | sentence | word list | story |
| ASR.1 | .268 | .464 | .458 | .365 | .285 | .464 | .458 | .368 |
| ASR.2 | .200 | .452 | .442 | .502 | .238 | .474 | .444 | .514 |
| ASR.3 | .325 | .559 | .549 | .537 | .407 | .588 | .578 | .571 |

**Table 4.** Cohen's Kappa and MCC values of three ASR systems after removing isolated words.

| Systems | Cohen's Kappa | MCC |
|---|---|---|
| ASR.1 | .437 | .438 |
| ASR.2 | .468 | .478 |
| ASR.3 | .555 | .585 |

### 5.3  Performance on Function and Content Words

Function words were separated from content words using NLTK [22]. We calculated agreement for function words and content words respectively with different measures (see Table 5). ASR.1 shows similar performance on function words and content words with slightly higher agreement for function words. Contrary to expectations, model ASR.2 and model ASR.3 present higher agreement for function words than content words.

**Table 5.** Cohen's Kappa and MCC values of three systems for function and content words.

|       | Cohen's Kappa | | MCC | |
|---|---|---|---|---|
|       | function | content | function | content |
| ASR.1 | .397 | .417 | .402 | .428 |
| ASR.2 | .510 | .453 | .533 | .465 |
| ASR.3 | .565 | .534 | .595 | .567 |

## 6  Discussion and Conclusions

In the current paper we investigated the performance of three ASR systems: the original ASR system (ASR.1) implemented in the reading tutor software used by first-graders, and the two newly developed ASR systems (ASR.2 and ASR.3) trained with JASMIN child speech data. We first obtained binary scores from the outputs of ASR systems and human transcripts as the reference. Then we applied different measures to evaluate the ASR performance.

The results show that with respect to Cohen's Kappa and MCC values, ASR.2 and ASR.3 present better performance than ASR.1. ASR.3 outperformed the other two systems. However, this is not always the case for all measures.

The F-measures of CA are similar for the three ASR systems. The precision of CA is higher for ASR.2 and ASR.3 while the recall of CA is lower for ASR.2 and ASR.3 compared to ASR.1. By using new ASR systems, the F-measure of CR was improved. Although there is a small difference in precision of CR among the three ASR systems, recall of CR significantly improved for ASR.2 and ASR.3. This suggests that lowering the false rejection rate (FRR) is important for further experiments because it influences precision of CR and recall of CA. It is notable that all three systems demonstrate relatively poor performance for Precision of CR, an important point that definitely deserves attention in future experiments.

We also saw different agreement performance on the four reading tasks implemented in the reading tutor. The two newly developed ASR systems achieved better results on sentences, word lists and stories than on isolated words. This is also a point that deserves attention. An important difference between the evaluation of isolated words by the annotator and the ASR system is that the annotator knows which words the pupils should be reading and can therefore compare the incoming signal to what she expected to hear. As a matter of fact, in a preliminary analysis in which we asked five elementary school teachers to give correct/incorrect judgements on 300 isolated words, we observed a very high agreement of 81.67% among the teachers, where all five teachers gave the same correct/incorrect judgement. At the moment, our ASR system does not know which word to expect and therefore cannot make such a specific form of comparison. Adopting a customized language model for isolated words would be a way of dealing with this discrepancy that is expected to enhance the performance of the system on isolated words.

The results demonstrated that ASR.2 and ASR.3 have better performance on function words than on content words while for ASR.1, better results were observed for content words. The expectation was that the ASR systems would provide more accurate feedback on content words than on function words. This was based on the assumption that the pupils would read content words more slowly and carefully. A possible explanation could be that function words occur more frequently and are more familiar to pupils. While content words, particularly new words, are often problematic for first-graders.

Although the present study was limited due to the small amount of transcribed speech data, the results are promising. More data collected from the pupils who used the reading tutor are being transcribed. Future directions for improving the acoustic models and the general performance of the ASR systems includes training with data not only from the JASMIN corpus, but also with a significant amount of new child speech data that are being collected and transcribed. We are also considering to apply transfer learning [13] to other sources of speech data, such as Dutch adult speech and Flemish (a variety of Dutch spoken in Belgium) child speech. Finally, we are examining other techniques for binary classification as a complement to ASR output feedback for children while reading, such as forced alignment [16].

## 7  Acknowledgements

The current research is carried out within the 'Dutch ASR-based Reading Tutor' (DART) project (http://hstrik.ruhosting.nl/DART/). This work is part of the Netherlands Initiative for Education Research (NRO) with project number 40.5.18540.121, which is financed by the Dutch Research Council (NWO). We would like to thank children who used the reading tutor at home during the pandemic, their parents and teachers who gave us informative feedback and advice in questionnaires and interviews and schools that participated in the experiments.

## References

1. Abdou, S.M., Hamid, S.E., Rashwan, M., Samir, A., Abdel-Hamid, O., Shahin, M., Nazih, W.: Computer Aided Pronunciation Learning System Using Speech Recognition Techniques. In: Proc. Interspeech. pp. 849–852. Pittsburgh, PA, USA, Sep. 17–21 (2006)
2. Bai, Y., Hubers, F., Cucchiarini, C., Strik, H.: ASR-Based Evaluation and Feedback for Individualized Reading Practice. In: Proc. Interspeech. pp. 3870–3874 (2020). https://doi.org/10.21437/Interspeech.2020-2842
3. Bai, Y., Hubers, F., Cucchiarini, C., Strik, H.: An ASR-based Reading Tutor for Practicing Reading Skills in the First Grade: Improving Performance through Threshold Adjustment. In: Proc. IberSPEECH 2021. pp. 11–15 (2021). https://doi.org/10.21437/IberSPEECH.2021-3
4. Banerjee, S., Beck, J., Mostow, J.: Evaluating the effect of predicting oral reading miscues. In: Proc. Interspeech. pp. 3165–3168. Geneva, Switzerland, Sep. 1–4 (2003)
5. Black, M.P., Tepperman, J., Narayanan, S.S.: Automatic prediction of children's reading ability for high-level literacy assessment. IEEE Transactions on Audio, Speech and Language Processing **19**(4), 1015 – 1028 (2011). https://doi.org/doi:10.1109/TASL.2010.2076389
6. Boughorbel, S., Jarray, F., El-Anbari, M.: Optimal classifier for imbalanced data using matthews correlation coefficient metric. Plos One **12**(6) (2017). https://doi.org/10.1371/journal.pone.0177678
7. Castles, A., Rastle, K., Nation, K.: Ending the reading wars: Reading acquisition from novice to expert. Psychological Science in the Public Interest **19**(1), 5–51 (2018). https://doi.org/10.1177/1529100618772271
8. Cucchiarini, C., Van Hamme, H.: The JASMIN Speech Corpus: Recordings of Children, Non-natives and Elderly People, pp. 43–59. Springer, Berlin, Heidelberg (2013). https://doi.org/10.1007/978-3-642-30910-6_3
9. Duchateau, J., Kong, Y.O., Cleuren, L., Latacz, L., Roelens, J., Samir, A., Demuynck, K., Ghesquière, P., Verhelst, W., Van Hamme, H.: Developing a reading tutor: Design and evaluation of dedicated speech recognition and synthesis modules. Speech Communication **51**(10), 985–994 (2009). https://doi.org/10.1016/j.specom.2009.04.010
10. Goldwater, S., Jurafsky, D., Manning, C.D.: Which words are hard to recognize? Prosodic, lexical, and disfluency factors that increase speech recognition error rates. Speech Communication **52**(3), 181 (2010). https://doi.org/10.1016/j.specom.2009.10.001

An ASR-based tutor for learning to read 1111. Hagen, A., Pellom, B., Cole, R.: Children's speech recognition with application to interactive books and tutors. In: 2003 IEEE Workshop on Automatic Speech Recognition and Understanding (IEEE Cat. No.03EX721). pp. 186–191 (2003). https://doi.org/10.1109/ASRU.2003.1318426
12. Hsu, L.: An empirical examination of efl learners' perceptual learning styles and acceptance of asr-based computer-assisted pronunciation training. Computer Assisted Language Learning **29**(5), 881–900 (2016)
13. Joshi, V., Zhao, R., Mehta, R.R., Kumar, K., Li, J.: Transfer learning approaches for streaming end-to-end speech recognition system (2020)
14. Kipyatkova, I., Karpov, A.: DNN-Based Acoustic Modeling for Russian Speech Recognition Using Kaldi. In: Ronzhin, A., Potapova, R., Németh, G. (eds.) Speech Comput. pp. 246–253. Springer International Publishing, Cham (2016). https://doi.org/10.1007/978-3-319-43958-7_29
15. Kocharov, D.: Automatic alignment of phonetic transcriptions for russian. In: Ronzhin, A., Potapova, R., Delic, V. (eds.) Speech and Computer. pp. 123–128. Springer International Publishing, Cham (2014)
16. Kraljevski, I., Tan, Z.H., Bissiri, M.P.: Comparison of forced-alignment speech recognition and humans for generating reference vad. In: Proc. Interspeech. pp. 2937–2941. Dresden, Germany, Sep. 6–10 (2015)
17. Kuhn, M.R., Schwanenflugel, P.J., Meisinger, E.B., Levy, B.A., Rasinski, T.V.: Aligning theory and assessment of reading fluency: Automaticity, prosody, and definitions of fluency. Reading Research Quarterly **45**(2), 230–251 (2010). https://doi.org/10.1598/rrq.45.2.4
18. Landis, J.R., Koch, G.G.: The Measurement of Observer Agreement for Categorical Data. Biometrics **33**(1), 159 (1977). https://doi.org/10.2307/2529310
19. Lee, A., Kawahara, T., Shikano, K.: Julius—an open source real-time large vocabulary recognition engine. In: EUROSPEECH 2001. pp. 1691–1694 (2001)
20. Li, X.L., Deng, L., Ju, Y.C., Acero, A.: Automatic children's reading tutor on hand-held devices. In: Proc. InterSpeech. pp. 1733–1736. International Speech Communication Association, Brisbane, Australia, Sep. 22–26 (2008), https://www.microsoft.com/en-us/research/publication/automatic-childrens-reading-tutor-on-hand-held-devices/
21. Limonard, S., Cucchiarini, C., van Hout, R., Strik, H.: Analyzing Read Aloud Speech by Primary School Pupils: Insights for Research and Development. In: Proc. Interspeech. pp. 3710–3714. Shanghai, China, Oct. 25–29 (2020). https://doi.org/10.21437/Interspeech.2020-2804
22. Loper, E., Bird, S.: Nltk: The natural language toolkit. In: In Proceedings of the ACL Workshop on Effective Tools and Methodologies for Teaching Natural Language Processing and Computational Linguistics. Philadelphia: Association for Computational Linguistics (2002)
23. McAuliffe, M., Socolof, M., Mihuc, S., Wagner, M., Sonderegger, M.: Montreal forced aligner: Trainable text-speech alignment using kaldi. In: Proc. Interspeech. pp. 498–502. Stockholm, Sweden, Aug. 20–24 (2017). https://doi.org/10.21437/Interspeech.2017-1386
24. Mohri, M., Riley, M.: Weighted finite-state transducers in speech recognition (tutorial). In: Proc. ICSLP. Denver, Colorado, USA, Sep. 16–20 (2002)
25. Mostow, J.: Is ASR accurate enough for automated reading tutors, and how can we tell? In: Proc. Interspeech. pp. 837–840. Pittsburgh, PA, USA, Sep. 17–21 (2006)
26. Mostow, J., Nelson-Taylor, J., Beck, J.E.: Computer-Guided Oral Reading versus Independent Practice: Comparison of Sustained Silent Reading to an Automated

Reading Tutor That Listens. Journal of Educational Computing Research **49**(2), 249–276 (2013). https://doi.org/10.2190/EC.49.2.g
27. Mostow, J., Roth, S.F., Hauptmann, A.G., Kane, M.: A prototype reading coach that listens. In: Proc. AAAI. pp. 785–792. Seattle, Washington, WA, Aug.–Sep. 31–4 (1994)
28. Pikulski, J.J., Chard, D.J.: Fluency: Bridge between decoding and reading comprehension. The Reading Teacher **58**(6), 510–519 (2005), http://www.jstor.org/stable/20205516
29. Povey *et. al*, D.: The Kaldi speech recognition toolkit. In: Proc. ASRU. pp. 1–4. Waikoloa, Hawaii, HI, USA, Dec. 11–15 (2011)
30. Qian, Y., Evanini, K., Wang, X., Lee, C.M., Mulholland, M.D.: Bidirectional lstm-rnn for improving automated assessment of non-native children's speech. In: Proc. Interspeech. Stockholm, Sweden, Aug. 20–24 (2017). https://doi.org/10.21437/Interspeech.2017-1386
31. Rao, P., Swarup, P., Pasad, A., Tulsiani, H., Das, G.G.: Automatic assessment of reading with speech recognition technology. Copyright 2016 Asia-Pacific Society for Computers in Education All rights reserved. No part of this book may be reproduced, stored in a retrieval system, transmitted, in any forms or any means, without the prior permission of the Asia-Pacific Society for Computers in Education. ISBN 9789868473591 p. 1 (2016)
32. Reeder, K., Shapiro, J., Wakefield, J., D'Silva, R.: Speech Recognition Software Contributes to Reading Development for Young Learners of English. International Journal of Computer-Assisted Language Learning and Teaching **5**(3), 60–74 (aug 2015). https://doi.org/10.4018/ijcallt.2015070104
33. Sudhakara, S., Ramanathi, M.K., Yarra, C., Ghosh, P.K.: An Improved Goodness of Pronunciation (GoP) Measure for Pronunciation Evaluation with DNN-HMM System Considering HMM Transition Probabilities. In: Proc. Interspeech. pp. 954–958 (2019)
34. Tejedor-García, C., Cardeñoso-Payo, V., Escudero-Mancebo, D.: Performance Comparison of Specific and General-Purpose ASR Systems for Pronunciation Assessment of Japanese Learners of Spanish. In: Proc. IberSPEECH 2021. pp. 6–10 (2021). https://doi.org/10.21437/IberSPEECH.2021-2
35. Tepperman, J., Black, M., Price, P., Lee, S., Kazemzadeh, A., Gerosa, M., Heritage, M., Alwan, A., Narayanan, S.S.: A Bayesian network classifier for word-level reading assessment. In: Proc. Interspeech. pp. 2185–2188. ISCA, Antwerp, Belgium, Aug. 27–31 (2007), http://www.isca-speech.org/archive/interspeech_2007/i07_2185.html
36. Wise, B., Cole, R., Vuuren, S.V., Schwartz, S., Snyder, L., Ngampatipatpong, N., Tuantranont, J.: Learning to Read with a Virtual Tutor: Foundations to Literacy. In: Kinzer, C., Verhoeven, L. (eds.) Interactive Literacy Education: Facilitating literacy learning environments through technology. Lawrence Erlbaum, Mahwah, NJ (2005), http://citeseerx.ist.psu.edu/viewdoc/summary?doi=10.1.1.120.7734